\pdfoutput=1

\documentclass[11pt]{article}

\usepackage[]{emnlp2021}
\usepackage{amssymb}

\usepackage{times}
\usepackage{latexsym}
\usepackage{graphicx}
\usepackage{multirow}
\usepackage{arydshln}
\usepackage{svg}

\usepackage[T1]{fontenc}

\usepackage[utf8]{inputenc}

\usepackage{microtype}

\usepackage[colorinlistoftodos,textsize=tiny]{todonotes}

\title{Relation Extraction from Tables using Artificially Generated Metadata}

\author{Gaurav Singh \\
  Amazon \\
  \texttt{singhgrv@amazon.co.uk} \And
  Siffi Singh \\
  Amazon \\
  \texttt{siffis@amazon.co.uk}\\ \AND
  Joshua Wong \\ 
  Amazon \\
  \texttt{joshuax@amazon.co.uk} \\ \And
  Amir Saffari \\
  Amazon \\
  \texttt{amsafari@amazon.co.uk}\\
  }

\begin{document}
\maketitle
\begin{abstract}
Relation Extraction (RE) from tables is the task of identifying relations between pairs of columns of a table. Generally, RE models for this task require labelled tables for training. These labelled tables can also be generated artificially from a Knowledge Graph (KG), which makes the cost to acquire them much lower in comparison to manual annotations. However, unlike real tables, these synthetic tables lack associated metadata, such as, column-headers, captions, etc; this is because synthetic tables are created out of KGs that do not store such metadata.
Meanwhile, previous works have shown that metadata is important for accurate RE from tables. To address this issue, we propose methods to artificially create some of this metadata for synthetic tables. Afterward, we experiment with a BERT-based model, in line with recently published works, that takes as input a combination of proposed artificial metadata and table content. Our empirical results show that this leads to an improvement of 9\%-45\% in F1 score, in absolute terms, over 2 tabular datasets. 
\end{abstract}

\section{Introduction}
Tables are a very useful tool in information representation because information can be stored and presented more concisely in a table compared to free text. Due to their ease and usefulness, a large number of tables are being produced and made available on the web \cite{zhang2020web} everyday. But tables are not just restricted to web, in fact, a lot of data across numerous domains is stored in tabular format. Not surprisingly then tables are a very large source of knowledge \cite{conf/www/LehmbergRMB16} for many real-world tasks.

Luckily, information contained in a table already has some implicit structure; there is a leading entity in every row, called the subject entity, and all other cells in the row are connected to this leading entity via some relation. However, despite this implicit structure, the information can not be easily converted to knowledge, since we do not know relations between columns w.r.t. a specific ontology. Therefore, to ingest this knowledge into KG, we need to map relations between columns to relations defined in the KG ontology \cite{ritze2015matching,ritze2017matching}, classify the entity-type and perform entity-linking. In this work we focus on one of these tasks i.e. Relation Extraction. 

\begin{figure}
    \centering
    \includegraphics[scale=0.22]{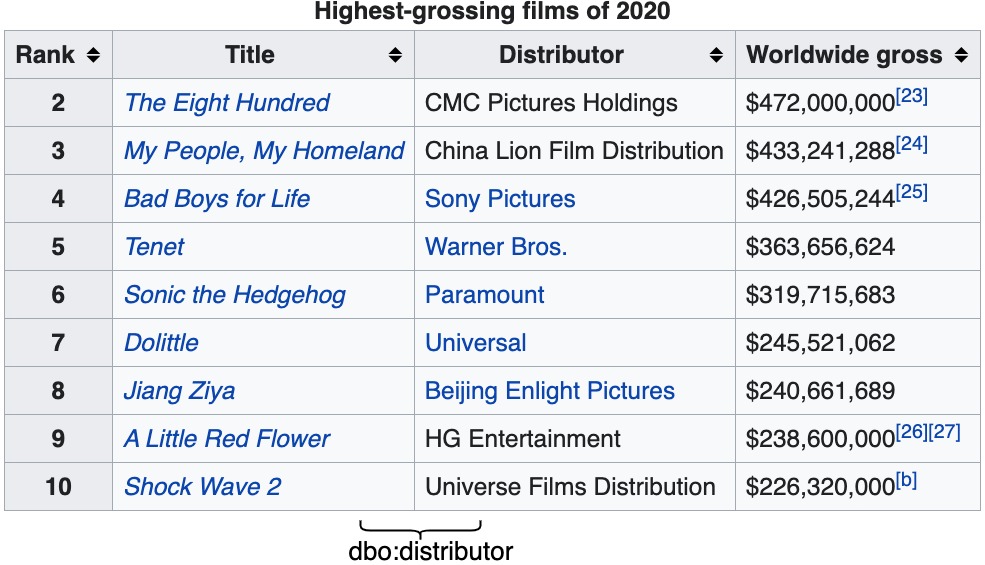}
    \caption{An example of relation extraction from tables. Here \textit{dbo:distribution} is the dbpedia relation between \textit{Title} and \textit{Distributor} columns.}
    \label{fig:re_example}
\end{figure}

In order to train a model for mapping relations between columns to KG relations, we require a dataset of tables labelled with KG relations. But, it would be very expensive to manually annotate tables for training such a model, and therefore, most current RE models are trained on labelled tables generated artificially from the KG \cite{jimenez2020semtab,cutrona2020tough}. To generate such a synthetic table, a fixed number (say, 10) of facts triples corresponding to a specific relation are repetitively sampled from the KG, and these are then used to generate 2-column tables, such that the entities on the left of relation form the left-column, and entities on right form the right-column. 

However, these synthetic tables usually lack two crucial pieces of information  that can provide useful signals for RE, namely: 1) context;  2) column-headers. \textbf{First} is a piece of text connecting different entities in the table. More specifically, an unstructured piece of text that implicitly or explicitly describes how two values in the same row are related. For instance, the sentence: "\textit{Paris} is the capital city of \textit{France}", connects \textit{Paris} and \textit{France} by the relation \textit{capital}. In real tables this information might be contained in the text present before or after the table, or in the captions, but since this context is not contained in KGs, it is also absent from synthetic tables created from KGs. \textbf{Second} is the headers for the columns of these synthetic tables. Most KGs do not contain information about possible column-headers for entities, and therefore, synthetic tables created from KGs do not contain column-headers as well. We should mention that KGs do contain information about entity-types, but these are conceptually different from column-headers, and can not be used as their replacement.  

Previously, there have been attempts \cite{deng2020turl, yin2020tabert} to resolve these issues by using real tables with metadata, whose entities are already linked to some public KG, such as DBpedia, as the training set. Unfortunately, such an approach does not work when the predictions are required w.r.t. a private or specific KG, as it would require extensive annotations. Another approach \cite{wang2021tcn} creates one-to-one mappings between relations of DBpedia and the target KG, and uses that mapping to convert relation predictions over DBpedia to target KG. However, such mappings require manual effort from trained annotators, and consequently, they are expensive to create. Also, they are often impractical due to significant ontological differences between the two KGs.

To address these issues, we propose two techniques to artificially create context and column-headers for synthetic tables generated from an arbitrary KG. Afterward, we experiment with a neural model that takes as input the table and artificially created metadata for relation prediction. Our model is in line with  recently published works \cite{yin2020tabert,deng2020turl} on modelling tabular data structures.  We perform experiments over 2 tabular datasets, consisting of 1 public benchmark dataset and 1 private dataset. Our empirical results show that the proposed method leads to very large improvements in RE performance over both datasets. Our contributions can be summarized as follows:
\begin{itemize}
    \item Propose methods to generate context \& column-headers for synthetic tables
    \item Show that proposed artificial metadata leads to large improvements in RE performance
\end{itemize}

\section{Methods}
\subsection{Generating Synthetic Tables}\label{gsm}
\begin{figure}
    \centering
    \includegraphics[scale=0.15]{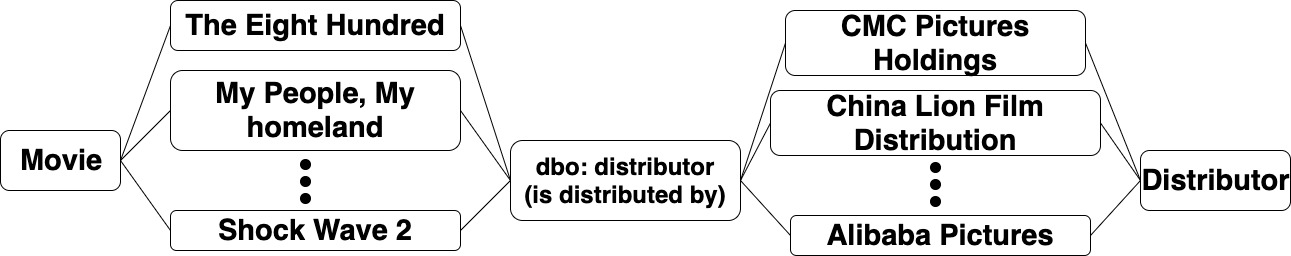}
    \caption{An illustration of a sub-graph that
represents the equivalent of the table in Fig \ref{fig:re_example}}
    \label{fig:subgraph_example}
\end{figure}

Tables can be viewed as sub-graphs of a KG, with nodes as the denotations of entities or literals. For example, the table from Fig \ref{fig:re_example} can be viewed as the sub-graph in Fig \ref{fig:subgraph_example}. To generate synthetic tables, we retrieve all sub-graphs with target relations. Afterward, we convert these sub-graphs to fact triples of the form: $(entity_1, relation, entity_2)$ and group them by relations. Thereafter, we first randomly draw the number of rows $R$ from the interval $[5, 10]$ to simulate variable size of real web-tables. We then select $R$ triples without replacement from the set of triples for that relation. Finally, we create $2$-column $R$-row tables using these $R$ triples, such that the left-entities from the first column, and the right-entities form the second column. 

To generate tables marked with the negative relation i.e. a special relation that denotes no relation or unknown relation, we pick left-column of a randomly chosen table, and combine it with the right-column of another randomly chosen table, to form a synthetic table. 

Unlike multi-column real tables, we create $2$-column synthetic table for 2 reasons: 1) relation between two columns is largely independent of other columns, 2) non-trivial to generate multi-column synthetic tables.

\begin{figure*}[t!]
    \centering
    \includegraphics[scale=0.45]{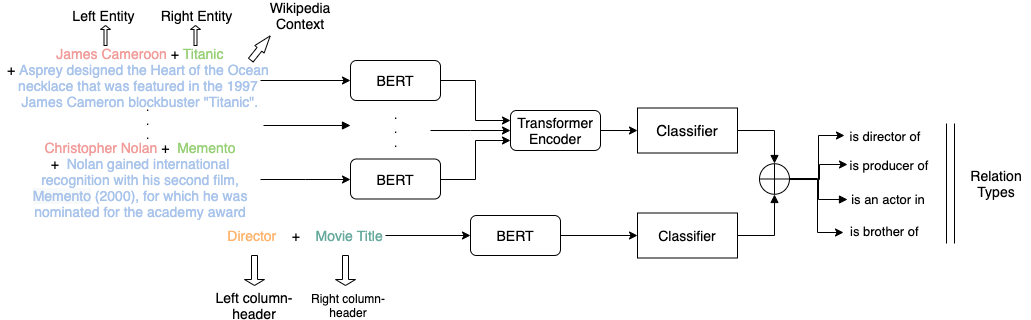}
    \caption{A schematic representation of the model architecture. The model takes as input column-pair, context, and headers to predict relation-type.}
    \label{fig:model_arch}
\end{figure*}

\subsection{Generating Metadata}\label{gm}
\paragraph{Context} We download the entire corpus of full-text articles from English language Wikipedia and cleanup the text files of any HTML tags and blank spaces. Afterward, we index all the paragraphs in the corpus using an Elastic-Search (ES) cluster. 

To retrieve context for a given pair of entities $(e_1,e_2)$, we first perform a logical AND query on the ES cluster to retrieve all the paragraphs that contain a mention of the two entities. If more than 10 such paragraphs are retrieved then we select the top-10 based on the score from ES cluster. We filter out all the returned paragraphs that have a length less than the added lengths of the two entities plus a constant i.e. $len(e_1)+len(e_2)+3$; this is to remove short noisy contexts that do not establish relation between $e_1$ and $e_2$. Thereafter, we select the top scoring paragraph and search for the topmost  sentence within it that mentions both $e_1$ and $e_2$. If we find such a sentence then we select it as the final context; otherwise, we select the topmost sentence that mentions $e_1$ and similarly, topmost sentence that mentions $e_2$, and concatenate the two sentences to form the final context.

\paragraph{Column-headers}
We randomly sample 10\% of English relational tables in the WDC dataset \cite{conf/www/LehmbergRMB16} i.e. 5 million tables. We use this sampled set to create an entity-to-header mapping between each entity and the header of the column in which that entity appears using a NoSQL database. To generate header of a column, we extract the most frequent headers for that entity in the entity-to-header table. We repeat this for all entities in that column, and collect one potential header candidate for each row. Afterward, we select the most frequent header across all rows in that column, and in case of a tie for the top place, we randomly select one of the top headers. 

         


\subsection{Model Architecture}

Our proposed model takes as input: 1) a column-pair; 2) wikipedia context for entity-pairs; 3) column-headers, and outputs: 1) relation-type (Fig \ref{fig:model_arch}). 

We linearize each row by joining the two cells and corresponding context using the [SEP] token. We prepend a [CLS] token to the linearized row, tokenize it using BERT subword tokenizer, and then pass it to a pretrained BERT \cite{devlin-etal-2019-bert} model. We take the vector representation of the [CLS] token as the representation of the row. We then pass the representation for all rows through a Transformer encoder to get a vector representation for the entire table. The transformer encoder layer performs attention induced averaging over rows to reduce noise in the entities and retrieved contexts.  Afterward, we pass this through a fully-connected layer to get prediction scores over all relations.

In parallel, we linearize column headers by joining the two headers using the [SEP] token. We prepend the [CLS] token to the linearized header row. Afterward, we pass this through a pretrained BERT encoder to obtain a vector representation for the combined headers. We then pass this representation through a fully-connected layer to get prediction scores over all relations. 

Finally, we add the scores obtained from rows and headers to get final prediction scores. At the end we choose the relation with the highest score as the predicted relation-type for the column-pair.

\section{Experimentation}
\subsection{Dataset}

\paragraph{Public dataset}
We evaluate our model on the public benchmark T2Dv2 \cite{chen2019colnet} dataset. It contains manually annotated mappings between column-pairs in Web-tables and relations in DBpedia KG. It consists of 779 tables, 89 relations, and 618 columns-to-relations annotations. Our train set consists of 2.2MM synthetic tables generated from the DBpedia KG using the method mentioned in Section \ref{gsm}. It consists of 659 relations in total, including a special "negative" relation to denote no-or-unknown relations.

\paragraph{Private Dataset}
This is a test set of internal catalog (IC) tables. This dataset consists of 5 tables, 65 annotated columns, and covers a set of 30 relations, including a special "negative" relation denoting no-or-unknown relations. For training set, we randomly sample a subset of facts belonging to these 30 relations from our internal knowledge graph (IKG), and use these to generate a set of 220K synthetic tables with $\approx$ 5-10 rows per table.

\subsection{Experimental Settings}
We use pretrained uncased BERT  (12-layer, 768-hidden, 12-heads, 110 MM params), from huggingface\footnote{https://github.com/huggingface} library. We include a dropout layer with a value of 0.5 before our classification layer and use the Adam optimizer, with a learning rate of $3e-5$, $\beta_1 = 0.9$, $\beta_2 = 0.999$ and $\epsilon = 1e-8$. We use a batch size of 16 and max sequence length of 256 to train the model. Hyper-parameters, i.e. epochs (values: $[1, 5]$)  and input sequence length (values: $[128, 256, 512]$) for BERT, were tuned on a validation set. The model takes $\approx$ 24 and 2.4 hours to complete 1 epoch on DBpedia and IKG respectively, using 4 Nvidia Tesla V100 GPUs.


\subsection{Results}
\begin{table}[t]
    \centering
    \begin{tabular}{l |c c c}
         Method & Pr & Re & F1 \\ \hline
         T2K  & 0.77 & 0.65 &  0.70\\ 
         A2P  & 0.70 & 0.84 & 0.77\\ \cdashline{1-4}
         Our & 0.65 & 0.78 & 0.71\\ 
    \end{tabular}
    \caption{Performance of different methods on T2Dv2 dataset. Results for T2K \cite{ritze2015matching} and A2P \cite{ritze2017matching} were copied from the papers. Note, results for T2K are on an older version of the dataset called T2D. All the metrics are micro in nature. 
    }
    \label{tab:result}
\end{table}

\begin{table}[t]
\begin{center}

    \begin{tabular}{l | l | c c c }
    
        Dataset & Input & Pr & Re &  F1  \\ \hline
        \multirow{3}{*}{T2Dv2} & T & 0.65 & 0.60 & 0.62\\ \cdashline{2-5}
                               & T + C & 0.57 & 0.80 & 0.67\\
                               & T + H & 0.60 & 0.73 & 0.66 \\
                               & T + C + H & 0.65 & 0.78 & 0.71\\ \hline
        \multirow{3}{*}{IC} & T  & 0.03 & 0.03 & 0.03 \\ \cdashline{2-5}
                               & T + C & 0.26 & 0.88 & 0.40 \\
                               & T + H & 0.17 & 0.54 & 0.26 \\
                               & T + C + H & 0.33 & 0.88 & 0.48 \\ \hline
    \end{tabular}
    \caption{Precision, recall and F1 score for our model after adding metadata.  Here, T, C and H stand for Table, Context and Headers respectively. All metrics are micro in nature.
    }
    \label{tab:ablation}
\end{center}    
\end{table}

Our performance is better than T2K \cite{ritze2015matching} in terms of F1 score (see Table \ref{tab:result}). We should mention that T2K is based on entity lookup in DBpedia, and therefore, by design can only work on tables that contain overlapping facts with DBpedia KG. In comparison, our method can work with tables that do not overlap with the target KG.

Similarly, A2P \cite{ritze2017matching} relies on building an attribute-to-property dictionary using T2K, and using it to match columns with property labels in the DBpedia KG. Consequently, it suffers from the same limitations as T2K. But it also makes a few additional assumptions that are not valid for all KGs and tables, such as, presence of natural language labels for relations in KG and column-headers in tables. It also relies on ad-hoc methods to reduce noise in the attribute-to-property dictionary, which may not easily generalize to other KGs, tables and domains. Our method does not make any of these assumptions; therefore, it can be generalized to various KGs and tables. 

\subsection{Ablation Study}
To understand the effects of proposed metadata on model performance, we perform experiments on both datasets by progressively including different metadata as model input (see Table \ref{tab:ablation}). We observe that separate inclusion of context and header leads to a significant improvement in F1 score over both datasets. We also observe that combined inclusion of context and header leads to the best performance. These results demonstrate the usefulness of proposed synthetic metadata in model performance.

To directly evaluate the quality of retrieved metadata, we performed manual analysis on a randomly sampled set of 100 synthetic tables. We observed that roughly 30\% of retrieved contexts in a table either, implicitly or explicitly, encode the correct relation between two columns. We also observed that roughly 60\% of retrieved column-headers were correct, while the rest were incorrect or blank.

\section{Conclusion and Future Work}
We propose general techniques to artificially generate useful metadata, specifically context and column-headers, for synthetic tables. By design, our proposed techniques can be applied on synthetic tables generated from any arbitrary KG. Our experiments on 2 tabular datasets show that such synthetic metadata leads to significant improvements in RE performance. 

In the future, we will explore following 3 research directions: 1) encode tables using pretrained table encoders for RE; 2) develop new methods to extract additional metadata for synthetic tables; 3) use natural language labels for relations in KG as input for model prediction.

\bibliography{anthology,custom}
\bibliographystyle{acl_natbib}

\appendix

\end{document}